\pdfoutput=1

\documentclass[11pt]{article}

\usepackage[]{acl}
\usepackage[utf8]{inputenc}

\usepackage{times}
\usepackage{latexsym}
\usepackage{graphicx}
\usepackage[justification=centering]{caption}

\usepackage[T1]{fontenc}

\usepackage[utf8]{inputenc}

\usepackage{microtype}

\title{Predicting Punctuation in Ancient Chinese Texts: A Multi-Layered LSTM and Attention-Based Approach}

\author{
  Tracy Cai\footnotemark[1] \\
  \texttt{cpcai@stanford.edu}
  \And
  Kimmy Chang\footnotemark[1] \\
  \texttt{kchang08@stanford.edu} \\
  \And
  Fahad Nabi\thanks{Equal contribution.} \\
  \texttt{nabif@stanford.edu} \\
}

\begin{document}

\maketitle

\section*{\hspace*{32mm}Abstract}

It was only until the 20th century when the Chinese language began using punctuation. In fact, many ancient Chinese texts contain thousands of lines with no distinct punctuation marks or delimiters in sight. The lack of punctuation in such texts makes it difficult for humans to identify when there pauses or breaks between particular phrases and understand the semantic meaning of the written text (Mogahed, 2012). As a result, unless one was educated in the ancient time period, many readers of ancient Chinese would have significantly different interpretations of the texts. We propose an approach to predict the location (and type) of punctuation in ancient Chinese texts that extends the work of Oh et al (2017) by leveraging a bidirectional multi-layered LSTM with a multi-head attention mechanism as inspired by Luong et al.'s (2015) discussion of attention-based architectures. We find that the use of multi-layered LSTMs and multi-head attention significantly outperforms RNNs that don't incorporate such components when evaluating ancient Chinese texts. 

\section{Introduction}
It was only up until the 20th century when Chinese adopted a proper punctuation system to denote pauses and line breaks between sentences. As a result, ancient Chinese contains thousands of lines with no distinct punctuation marks present in its script. If one isn't educated to read such style of text,   they are unlikely to decipher the writer's intended meaning of the content they are reading (i.e., poems, newspapers, journals, etc). Instead, they may hold on to multiple distinct interpretations of the written texts. This means only experts in ancient Chinese who can read the script can understand the meaning of such traditional texts. To make ancient Chinese texts decipherable to today's readers, the placement of punctuation is crucial to understanding the true meaning of written text (Mogahed, 2012).

Previous approaches have experimented with Encoder-Decoder RNNs, GRU, and LSTMs as well as different single-headed attention structures (local and global) to successfully conduct language translation tasks. One recent work that built an efficient model for optimal performance in a task similar to ours (predicting line breaks) is that of Oh et al. In Oh et al (2017), researchers were able to predict where line breaks ought to be in Hanmun (a punctuation-lacking Korean script) with a multi-layered LSTM model that incorporated an end-of-sentence attention mechanism. As Luong et al. (2015) found local attention models to significantly outperform non-attentional ones on translation tasks between English-German, we were inspired to improve upon Oh et al.'s approach towards line-break prediction by paying special attention to the attention model. 

While Oh et al.'s multi-layered attention-focused LSTM model was efficient in predicting line breaks, we aim to improve upon their work in three ways. More specifically, we curate a more diverse set of training data (poems from various time periods in the ancient era which include rare and unique words), leverage a multi-head attention mechanism, and go beyond predicting line-breaks by including the exact location (and type) of punctuation in ancient Chinese texts. 
\section{Prior Literature} 

There have been many exciting findings in prior literature that are relevant to our goal for predicting punctuation markers in ancient Chinese. 

One finding is that multi-layered long short-term memory models (LSTM) offer high performance and accuracy when analyzing ancient poetry (Ghosh et al. and Oh et al.) as compared to RNN and GRU counterparts. In Ghosh et al, LSTM was used to analyze stroke features of Devanagari and Bengali because previous papers' use of Hidden Markov Models struggled to segment cursive handwritten texts into basic features resulting in low recognition rates. Unlike HMMs, Bidirectional LSTMs (two-layered) are contextually aware (both in the past and future) of the sentence due to the availability of two hidden layers for processing. They are able to then recognize unknown words (not present in training) as the model recognizes the strokes present in the unknown word and classifies it to a particular stroke class. Thus, we have greater recognition accuracy and higher performance with BLSTMs. Similarly, in Oh et al, they found that LSTM models exhibited the best precision, recall, and F1-scores relative to other RNN and GRU models when predicting where sentences should be segmented. By stacking one, two, and three additional layers, they found that the LSTM model's respective scores increased in each category.

Another finding is that the presence of rare or unique words and phrases are dominant challenges in analyzing ancient Chinese poetry (Zhang et al, Koehn et al. and Oh et al.). Zhang et al's \emph{cold start} approach for expanding knowledge discovery of emotional words in ancient poetry (integrating an emotional word into a poetry dictionary, matching it to a theme, and generating a learning corpus from it) was flawed when faced with rare words which may convey intense emotions but are not likely used under a particular theme. This aligns with Koehn et al's analysis of infrequent words being difficult to translate by NMT systems (namely, adjectives and verbs) and Oh et al's lackluster performance in predicting line breaks when faced with rarely-used characters in the training data. 

The last finding is that incorporating an attention mechanism to a model tends to improve its performance as the model learns to place greater value towards relevant portions of a given input. Just as Luong et al found local attention models to outperform non-attentional ones on translations between English-German, Oh et al found their best performance in line-break tasks to be driven by a multi-layered LSTM model that was trained with an attention mechanism (to focus on the end-word while introducing line breaks). It is noteworthy that although the sub-tasks in which attention is used are different (translation tasks and line-break segmentation), incorporating some kind of local and/or global attention seemed to make the model more contextually aware and accurate. One key limitation that Oh et al. face is that SMTs seemed to offer greater translation quality than their attention-based NMT on sentences with more than 60 subword tokens, whose quality sharply declined over 80 subword tokens. We aim to improve upon Oh et al.'s existing attention model by incorporating a multi-head attention mechanism. 

\section{Data}
The data source for our project is the Chinese Poetry dataset on Github:
\begin{itemize}
\raggedright
    \item \url{https://github.com/chinese-poetry/chinese-poetry}
\end{itemize}

It contains a hierarchy of folders sorted by author and/or grouping of subjects. The collection features a diverse set of writings that are not limited to poetry including analects, stories, phrases, philosophies, etc. Our reasoning for using a (primarily) poetry dataset is that ancient Chinese texts are generally lyrical and poetical.

Using the json files from this Github, we created a dataset for our project. Our final dataset consists of 341,531 data points and eight features. Each data point represents a line in the text (i.e. one item in a given array of text that represents a stanza/paragraph). The eight features are title (string), author (string), genre (string), stanza number/line number (integer), punctuation indices (list), punctuation points (list), the original text (string), and the input text (string). The 12 punctuation points accounted for are available in the Appendix (see Figure \ref{fig:punc}). Genre was sorted into one of three categories: poetry, teachings, and philosophy.

Below is a list of the specific writings used along with a short description:
\begin{itemize}
    \item \textit{caocao.json (293 data points)} - poetry. Cao Cao was a warlord and accomplished poet who lived in the final years of the Eastern Han dynasty who lived in the years 155-220 CE.
    
    \item \textit{chuci.json (2273 data points)} - poetry. Chu Ci or the Songs of Chu is an ancient anthology of Chinese Poetry from the Han dynasty (around 221 BC).
    
    \item \textit{ci folder (159,049 data points)} - more excerpts from Songs of Chu (multiple json files), poetry.
    
    \item \textit{lunyu.json (512 data points)} - philosophy. Lunyu, or the ``Confucian Analects," is a collection of Confucian sayings and dialogues that has served as a source of the philosophies of Confucius. It was compiled around the years (500 BC-8 CE).
    
    \item \textit{dizigui.json (90 data points)} - teachings.
    Di Zi Gui or ``Standards for being a Good Pupil and Child" was written in the Qing dynasty under the Kangxi Emperor (1661-1722 CE) by Li Yuxiu. Note: here punctuation was determined with spaces as well.
    
    \item \textit{guwenguanzhi.json (1016 data points)} - teachings. Guwen Guanzhi is an anthology of essays published during the Qing dynasty in 1695.
    
    \item \textit{qianjiashi.json (614 data points)} - poetry. The Qianjiashi or ``Poems of one thousand Writers" is an anthology of poems from the Tang (618-907 CE), Five dynasties (907-960 CE), and Song periods (960-1279 CE). 

    \raggedright
    \item \textit{sanzijing-traditional.json (96 data points)} - teachings. San Zi Jing or the ``Three-Character Classic" was written in the 13th century CE during the Song dynasty. It served as the basis of elementary education during the Ming and Qing dynasties \textit{Johnson \& Nathan}.
    
    \item \textit{shenglvqimeng.json (90 data points)} - poetry. Sheng Lv Qi Meng or Signifances of Enlightenment. Unknown publication date.
    
    \item \textit{tangshisanbaishou.json (1531 data points)} - poetry. Tang Shi San Bai Shou or ``The Three Hundred Tang Poems" was compiled by the Qing scholar Sun Zhu and published in 1764 CE. 
    
    \item \textit{youxueqionglin.json (191 data points)} - philosophy. You Xue Qiong Lin was compiled in the Ming (1368-1644 CE) and Qing (1644-1911 CE) period.
    
    \item \textit{zengguangxianwen.json (752 data points)} - teachings. Zengguang xianwen or ``enlarged writings of worthies" was widely circulated in the late Qing period (1644-1911 CE).
    
    \item \textit{zhuzijiaxun.json (50 data points)} - teachings. Zhu Zi's family training was written during the Song Dynasty (960-1279 CE).
    
    \item \textit{quantangshi.json (125,504 data points)} - poetry. Quan Tangshi or ``Complete Tang Poems" is the largest collection of Tang poetry (the Tang dynasty ranged from 618 to 907 CE).
    
    \item \textit{shijing.json (1319 data points)} - poetry. Shiijng or ``Odes" is the oldest collection of Chinese poetry with works dating from the 11th to 7th centuries BC.
    
    \item \textit{daxue.json (15 data points)} - teachings. Daxue or ``Great Learning" is a Chinese text written around 551-479 BC by Confucius.
    
    \item \textit{mengzi.json (690 data points)} - philosophy. Collection of stories of the Confucian philosopher Meng Ke who lived in the 372-289 BC.
    
    \item \textit{zhongyong.json (38 data points)} - philosophy. Zongyong or the ``Doctrine of the Mean" is one of the Four Books of classical Chinese philosophy written by Zisi (481-402 BC).
    
    \item \textit{nantang.json (185 data points)} - poetry. Poetry from Nantang in the Tang dynasty (618-907 CE).
    
    \item \textit{huajianji.json (1711 data points)} - poetry. Collection of poems written by Huajian Ji originally published in 940 CE.
    
    \item \textit{yuanqu.json (45,509 data points)} - poetry. Poetry from the Yuan dynasty (1271-1368 CE).
 \end{itemize}

There are 338,078 data points in the poetry genre, 2021 data points in the teachings genre, and 1432 data points in the philosophy genre. Thus, the majority of the data points are from the poetry genre. The writings range from 11th century BC to 1911 CE.

\section{General Approach and Model} 

To effectively predict punctuation markers in ancient Chinese, we will build an Encoder Decoder Seq-to-Seq Model with Attention added to the encoder's output. The encoder and decoder each consist of LSTM layers.

\subsection{Long Short-Term Memory (LSTM)}
\begin{figure}[h]
    \centering
    \includegraphics[width=0.4\textwidth]{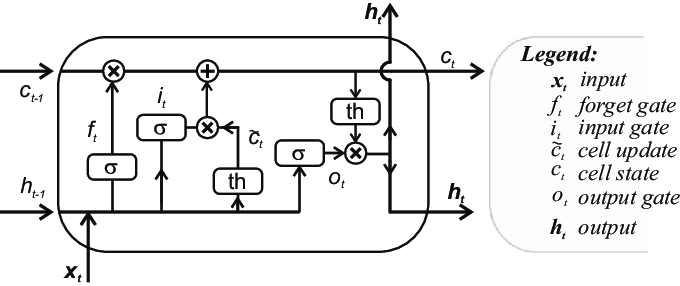}
    \caption{LSTM cell}
    \label{fig:LSTM_cell}
\end{figure}

The LSTM architecture serves to control the flow of information that is necessary for a prediction. In particular, the cells in a LSTM's hidden layers feature an input, output, and forget gate that determines what should be included and excluded from a certain state. Input and output gates open and close in order to let LSTM cells store information until the forget gate permits a cell to erase error information. A diagram of an LSTM cell is shown in Figure \ref{fig:LSTM_cell}.

We particularly chose the LSTM architecture as it had the best precision, recall, and an F1-score in Oh et al.'s (2017) approach (see Figure \ref{fig:oh} in the Appendix). 

When experimenting with LSTM architecture, Oh et al also found that as the number of layers increased by one, so did the performance of the LSTM architecture. By stacking three LSTM layers and incorporating the attention mechanism, their model became even more accurate with learning sequential data and predicting line breaks (see Figure \ref{fig:LSTM} in the Appendix).  

\begin{figure}[h]
    \centering
    \includegraphics[width=0.4\textwidth]{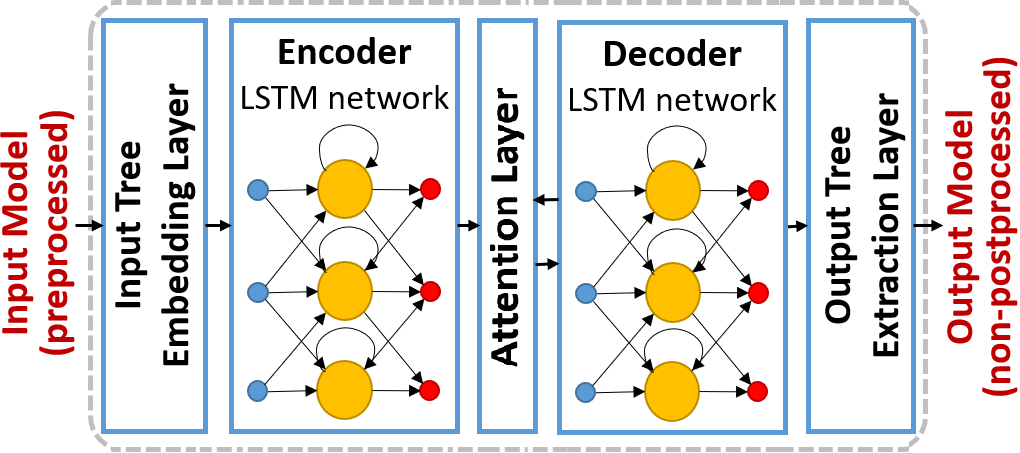}
    \caption{Core Model Architecture}
    \label{fig:core}
\end{figure}

For our core model, we aim to incorporate the attention mechanism for a three-layer stacked LSTM model. The model architecture is shown in Figure \ref{fig:core}. Along with a prediction $Y$, the attention mechanism will provide a context vector $C_i$ which contains the weighted average over a sequence of words in the data. We aim to explore both global (all of the hidden states in the encoder are considered when computing $C_i$) and local attention (which focuses on a subset of source positions for a given target word) mechanisms in our core model. We hypothesize that local attention will yield the best accuracy and performance as it improves contextual awareness (and has the possibility of paying close attention to the end-words of sentences). 

\subsection{Attention Mechanism}
Similar to \textit{Oh et. al} where the end word was important for the location of line breaks, we decided that words preceding punctuation points play a crucial role in the location of punctuation. Thus, we believe that by adding attention, we will be able to improve the performance of our model. 

Attention builds upon our core model, which involves an encoder-decoder system. Encoding entails processing the input by encoding it into a context vector that summarizes the input. Decoding is where the model translates the context vector into the desired output. The addition of attention occurs in the encoding stage. Attention adds focuses on the most relevant data so that the decoder can translate the data and make improved decisions. It can be seen as a vector of significance weights.

\begin{figure}[h]
    \centering
    \includegraphics[width=0.4\textwidth]{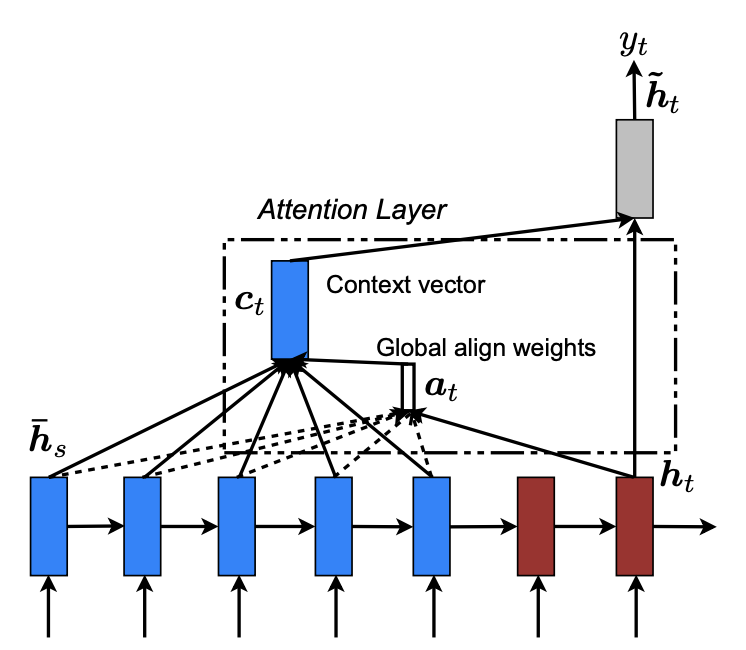}
    \caption{Global Attention}
    \label{fig:global}
\end{figure}

\textit{Luong et. al} classified attention-based models into global and local attention. Global attention considers all hidden states of an encoder when creating a context vector $\mathbf{c}_t$. At each time step, the model infers a \textit{variable-length} alignment weight vector $\mathbf{a}_t$ based on the current target state $\mathbf{h}_t$ and all source states $\bar{\mathbf{h}}_s$ (see Figure \ref{fig:global}). 

In contrast, the local attention model infers a \textit{variable-length} alignment weight vector $\mathbf{a}_t$ based on the current target state $\mathbf{h}_t$ and source states $\bar{\mathbf{h}}_s$ in a window centered around the source (see Figure \ref{fig:local} in the Appendix). The drawback of global attention is that it attends to all words, which is expensive for long sequences. Local attention avoids expensive computations, but has the trade-off of a reduced subset of source positions for each target character.

Along with exploring local and global attention, we aim to incorporate multi-head attention into our LSTM model (see Figure 7). A multi-head attention design has several attention layers running in parallel with one another. For example, consider the global attention layer from above. For each head (i.e., specified attention layer), we would linearly project the given queries, keys, and values using varying weight matrices. After feeding them into a parallel attention pool, we would have \emph{h} attention layer pooling outputs which would be concatenated and linearly transformed with another projection for a final input. 

As a result of multi-head attention, our multi-layered LSTM model would be able to learn information from various different representation
subspaces (at distinct positions) at the same time. The model would be able to attend to different parts of the input sequence based on the particular head. 

\section{Methods} 

\subsection{Metrics} 

The primary metric we intend to use is accuracy to evaluate both punctuation index and type prediction. 

For punctuation index, we will look at how many correct v. incorrect index predictions are made. Moreover, we will look at if there are punctuation types that are more commonly predicted accurately.

For type prediction, we will construct a confusion matrix and provide analysis on the precision and recall for each punctuation as well as macro and weighted F1 score. Precision will identify the proportion of punctuation predictions of a particular punctuation type are actually correct (see Equation \ref{eq:prec})\footnote{Note that TP, TN, FP, FN stand for true positive, true negative, false positive, ad false negative, respectively.}. 

\begin{equation} \label{eq:prec}
\textnormal{Precision} = \frac{TP}{TP + FP} 
\end{equation}

\noindent Recall will identify the proportion of a particular punctuation type are predicted correctly (see Equation \ref{eq:rec}). 

\begin{equation} \label{eq:rec}
\textnormal{Recall} = \frac{TP}{TP + FN} 
\end{equation} 
\noindent Weighted F1 score is calculated by taking the mean of all per-class F1 scores while considering each class's support (see Equation \ref{eq:wf1})\footnote{$N$ is the total number of classes and $n_i$ is the number of data points in a class.}. 

\begin{equation} \label{eq:wf1}
\textnormal{Weighted F1} = \frac{1}{N}\sum_{i=1}^N \left(\textnormal{Class i's F1} \cdot n_i\right)
\end{equation}

\noindent Thereby it is an indicator for precision and recall across the punctuation types (weighted accordingly). Macro F1 score simply takes the arithmetic mean across punctuation types and does not account for the difference in punctuation type occurrence (see Equation \ref{eq:mf1}). 

\begin{equation} \label{eq:mf1}
\textnormal{Macro F1} = \frac{1}{N}\sum_{i=1}^N \left(\textnormal{Class i's F1}\right)
\end{equation} 

\subsection{Input Datasets and Preprocessing}
From our datasets, we extracted and recorded all instances of punctuation and their respective indices in the poems. These punctuation marks were then replaced in the original text to create input data for our models.  

For example, our first data point is a a sentence from the first stanza of the original poem:

    \includegraphics[width=0.3\textwidth]{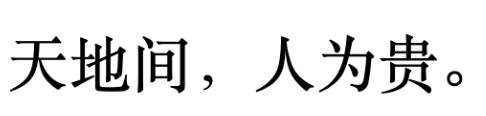} 

After preprocessing (parsing through the original sentence and extracting all relevant punctuation marks), this original sentence becomes our input for the model: 

\includegraphics[width=0.25\textwidth]{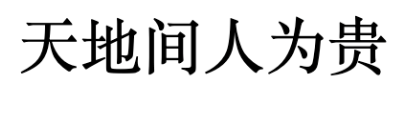} 

Key metadata that we preserve are the literal punctuation marks that are removed and their respective index numbers in the poem. In this example, our punctuation list is composed of the punctuation located at the respective indices [3, 7].A subset of what our preprocessed meta data looks like for our model is available in the Appendix (see Figure \ref{fig:metadata}). And, our key data that we feed as inputs into the model is also available in the Appendix (see Figure \ref{fig:input}):

As the final dataset consists of 341,531 data points, we aim to use approximately 80$\%$ of the dataset (273,225 input sentences) as training data and 20$\%$ of the dataset (68,306 input sentences) as testing data. The reason for this is to ensure our model is exposed to all types of punctuation marks.  

\subsection{Baseline Model}

\begin{figure}[h]
    \centering
    \includegraphics[width=0.4\textwidth]{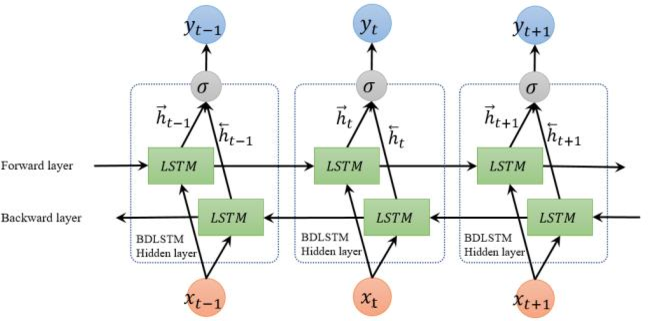}
    \caption{Baseline Model}
    \label{fig:base}
\end{figure}

Our baseline model is a bidirectional Recurrent Neural Network (BRNN) with Long Short-Term Memory (LSTM) cells (see Figure \ref{fig:base}). 

For a particular input $X$ (namely, the input sentence), the output $Y$ will be a string of indices (namely, the predicted indices of line-breaks separated by “/”). An example of the output prediction is for the very first data point is ``4/12/1/2/3/9/5/”. For this particular baseline model, we begin with mapping a sequence of ancient Chinese words to embedding vectors which we learn in the embedding layer. 

After the embedding layer, such input data is fed into our LSTM model followed by a hidden linear and classification layer for output predictions. The training size of our data is 180,000 and testing size of our data is 750. We aim to tune the development of our the model using different combinations of data point types (of all of the different genres and time periods these stories/poems were written in). 

\subsubsection{Recurrent Neural Networks (RNNs)}
RNNs pose the advantage of learning long-term dependencies without keeping redundant context information. Specifically, RNNs learn probability distributions over sequential data in order to predict the next symbol in a sequence. This is represented in Equation \ref{eq:rnn}, where $f$ is a softmax function that predicts the probabilities of a particular index containing any of the original punctuation marks. There are three layers in a RNN: input layer, hidden layer(s), and output layer. One major drawback of RNNs is the vanishing and/or exploding gradient problem. This happens when sequences get long and the model fails to train due to null/exploding weights. A solution to this is the use of an LSTM network.

\begin{equation} \label{eq:rnn}
g_t = f(g_{t-1}, x_t))
\end{equation}

\noindent BRNNs (in contrast to unidirectional RNNs) use data from both future and prior data in order to make predictions on the current state. 

\subsection{Core Model}
\begin{figure}[h]
    \centering
    \includegraphics[width=0.4\textwidth]{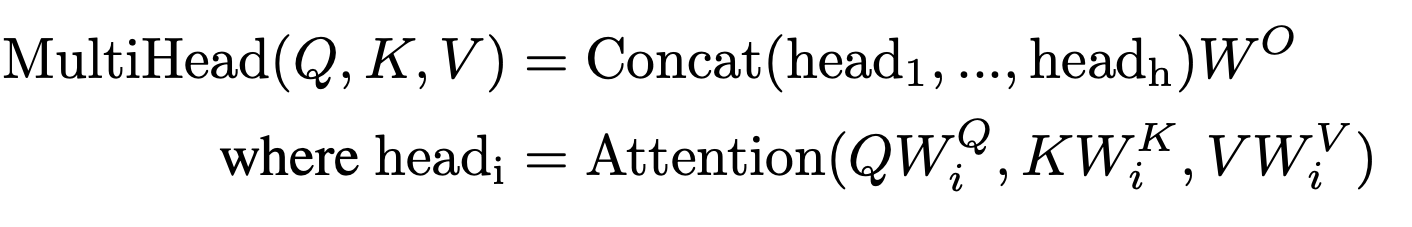}
    \caption{Multi-Head Attention}
    \label{fig:multihead}
\end{figure}

Our core model is a Encoder-Decoder Sequence to Sequence model where both the encoder and decoder have a three-layer stack of bidirectional LSTM cell units. We apply multi-head attention as an intermediate layer between the encoder and decoder. We define multi-head attention mechanism as see in Figure \ref{fig:multihead}.

After multi-head attention is applied to the output of the encoder, the decoder is able to learn from all of the positions in the input sequence while paying special attention to specific parts of the input model.

\section{Results and Analysis} 
Before delving into the results, we establish some baseline figures. The percentage of punctuation indices out of all indices is approximately 15.63\%. Thus, if our models were to guess no punctuation for all indices, then the baseline accuracy would be 15.63\%. Nevertheless, for the baseline and original model, we look to only predict two different classes: punctuation and no punctuation. Thus, the baseline accuracy that we chose to evaluate our model on is 50\%. For the original model, we also included the results for class-specific punctuation. Here, we exhibit four classes: comma, period, other punctuation, and no punctuation. The distribution across these classes are approximately 7.42\%, 7.37\%, 0.85\%, 84.37\%, respectively. We chose to focus on comma and period since they respectively account for 47.45\% and 47.11\% of the punctuation seen in the dataset. Thus, the other types of punctuation only account for 5.44\% of the full set of punctuation types. We establish the baseline accuracy to be 25\% for the  class-specific punctuation results.

Our baseline model (bidirectional RNN with LSTM cells) exhibited an accuracy of 0.171, precision of 0.086, recall of 0.230, and f1-score of 0.111. Thus, we conclude that the baseline model is unable to beat the baseline accuracy and that a different model is needed to approach the task of punctuation prediction.

We evaluated our original model (multi-layered LSTM with attention) based on whether it could predict (i) if a punctuation marker ought to be present at a particular index (of any type) and (ii) the correct type of punctuation at a particular index. For predicting whether a given punctuation ought to belong at a particular index, our results are found in Table \ref{table:nonlin1}.

\begin{table}[h]
\caption{General Punctuation Results} 
\centering 
\begin{tabular}{c c c c} 
\hline\hline 
Accuracy & Precision & Recall & F1-Score \\ [0.5ex] 
\hline 
0.813 & 0.180 & 0.004 & 0.009 \\ [1ex] 
\hline 
\end{tabular}
\label{table:nonlin1} 
\end{table}

For predicting the type of punctuation at a given location, we segment our results in four classes: no punctuation, comma, period, and other punctuation. Below is a summary of our results from each class (see Table Table \ref{table:nonlin2}), weighted scores (see Table Table \ref{table:nonlin3}), and macro scores (see Table \ref{table:nonlin4}). 

\begin{table}[h]
\caption{Class-Specific Punctuation Results} 
\centering 
\begin{tabular}{c c c} 
\hline\hline 
Class & Precision & Recall \\ [0.5ex] 
\hline 
No Punctuation & 0.835 & 0.983 \\
Comma & 0.092 & 0.006 \\
Period & 0.088 & 0.016 \\
Other Punctuation & 0 & 0 \\ [1ex] 
\hline 
\end{tabular}
\label{table:nonlin2} 
\end{table}

\begin{table}[h]
\caption{\textbf{Weighted} General Punctuation Results} 
\centering 
\begin{tabular}{c c c c} 
\hline\hline 
Accuracy & Precision & Recall & F1-Scores \\ [0.5ex] 
\hline 
0.822 & 0.711 & 0.822 & 0.757 \\ [1ex] 
\hline 
\end{tabular}
\label{table:nonlin3} 
\end{table}

\begin{table}[h]
\caption{\textbf{Macro} General Punctuation Results} 
\centering 
\begin{tabular}{c c c c} 
\hline\hline 
Accuracy & Precision & Recall & F1-Scores \\ [0.5ex] 
\hline 
0.822 & 0.254 & 0.251 & 0.235 \\ [1ex] 
\hline 
\end{tabular}
\label{table:nonlin4} 
\end{table}

The original model's accuracy of 81.3\% in the case of general punctuation far surpasses our baseline model and successfully tops the baseline accuracy of 50\%. Thus, it appears that the multi-layered LSTM model is adequate at predicting where punctuation markers belong in ancient Chinese scripts.

When comparing predictions for whether a punctuation marker should be present versus the particular type of punctuation, the multi-layered LSTM network was marginally more accurate (by 0.9\%) in predicting the latter and had higher weighted and macro-averaged precision/recall/f1-scores. In both situations, the original model was accurately able to identify whether a punctuation ought to be present (0.813) and which class it belongs to (0.822). Of note, the class-specific punctuation case far surpassed the baseline accuracy of 25\%.

Out of all the classes, the original model for general punctuation performed well with detecting locations where there should be no punctuation in the text. However, it seems to predict a lot of commas and periods in locations where there in fact weren't any. Similarly, it predicts that there are no commas or periods in indexes where there ought to be. More specifically for the general punctuation results, the 0.180 precision entails that 18\% of the predictions for punctuation are correct. The 0.004 recall entails that 0.4\% of the time that there is punctuation present at the index, the model predicts that there is punctuation there. For the case of class-specific punctuation, we see a similar model weakness. The low precision and recall for the comma and period classes show that the model is not able to predict the existence of a comma or period when there should be one at a given index. Notably, the model seemed to never predict other punctuation markers besides commas and periods that ought to be in the data.

One possible reason for the low precision and recall results is that we have imbalanced data within our punctuation markers classes (commas, periods, other punctuation). To remedy the high false positive/negative rates, we could try to over-sample such classes when preparing model data. To remedy the failure in predictions for other punctuation markers, we can dedicate a portion of the dataset as validation (with such markers present), tuning appropriate hyper-parameters without overfitting.  

Between the weighted and macro-averaged results when predicting punctuation markers, the weighted results were significantly higher for precision, recall, and f1-scores. This shows that precision, recall, and f1-scores must have dropped for classes (namely, commas and periods) that were underrepresented in the dataset as a whole. As it is underrepresented, it is weighed less towards the total precision, recall, and f1-scores. Although we were able to accurately pick up appropriate punctuation markers, we would want to ensure our model is trained with a class that has fair representation. 

Along with these results, we created a confusion matrix to analyze differences between predicted values and actual values per class. A breakdown of our prediction outcomes for whether there should be a punctuation marker are shown in Figure \ref{fig:num12} and the particular punctuation marker prediction by class are shown in Figure \ref{fig:num13}. In Figure \ref{fig:num12}, class 0 represents no punctuation and class 1 represents punctuation. In Figure \ref{fig:num13}, class 0 represents no punctuation, class 1 represents comma, class 2 represents period, and class 3 represents other punctuation.

The confusion matrices (see Appendix) substantiates the precision and recall findings for both general and class-specific punctuation. Indeed, for general punctuation, the model makes a considerable number of predictions that there is punctuation present where there is not. For class-specific punctuation, the model rarely predicts the existence of comma, period, and other punctuation (white spaces for labels 1, 2, and 3) and guesses a considerable number of comma, period, and other punctuation where there should be no punctuation.

\section{Conclusion} 

Building on the work of Oh et al. (2017) for predicting line breaks in Hanmun, we extended their approach to create a model that would generalize such predictions for punctuation markers and types in ancient Chinese. We did this by building a Encoder-Decoder Sequence to Sequence model with multi-layered stacked LSTM units and a multi-head attention mechanism between the encoded and decoded representations. We found that by incorporating multiple layers of an LSTM and a multi-head attention mechanism, we were able to make more accurate predictions for when punctuation markers were present or not. Although our model's performance could be improved for class-specific punctuation (commas, periods, other punctuation), we have a solid method for predicting whether or not punctuation markers belong at a particular index in an ancient script. 

\section{Future Work}
Our model currently surpasses performance expectations for predicting when there ought to be no punctuation markers at a given index (which fits the data given the ratio of characters to punctuation markers). However, a key project limitation is our subpar performance in recall and precision for predicting class-specific punctuation markers (namely, commas, periods, and other punctuation). We believe this is due to a class imbalance in the datasets for each respective class. For future work, we advocate to either randomly under-sample from the no-punctuation class or randomly over-sample from the commas/periods/other classes. Alternatively, we can tune different LSTM-based models such that they are unique for each class and combine them for improved performance. 

\newpage

\section{References}

\raggedright

Johnson, David; Andrew James Nathan (1987). \textit{Popular Culture in Late Imperial China}. University of California Press. p. 29. ISBN 9780520061729. \\[0.5em]

\vspace{\baselineskip}

Luong, Ming-Thang; Pham, Hieu; Manning, Christopher (2015). Effective Approaches to Attention-based Neural Machine Translation. Available at: \url{https://arxiv.org/pdf/1508.04025.pdf}. \\[0.5em]

\vspace{\baselineskip}

Mogahed, Mogahed M. (2012). Punctuation Marks Make a Difference in Translation: Practical Examples. Available at: \url{https://files.eric.ed.gov/fulltext/ED533736.pdf}. \\[0.5em]

\vspace{\baselineskip}

Oh, D. H.; Shah, Z.; Jang, G.-J. (2017). Line-break prediction of Hanmun text using recurrent neural networks. \textit{International Conference on Information and Communication Technology Convergence}. \\[0.5em]

\vspace{\baselineskip}

Popel, M.; Tomkova, M.; Tomek, J.; Kaiser, Ł.; Uszkoreit, J.; Bojar, O.; Žabokrtský, Z. (2020). Transforming machine translation: A deep learning system reaches news translation quality comparable to human professionals. \textit{Nature Communications}. \\[0.5em]

\raggedright

\section{Appendix}\label{sec:appendix}

\begin{figure}[h]
    \centering
    \includegraphics[width=0.35\textwidth]{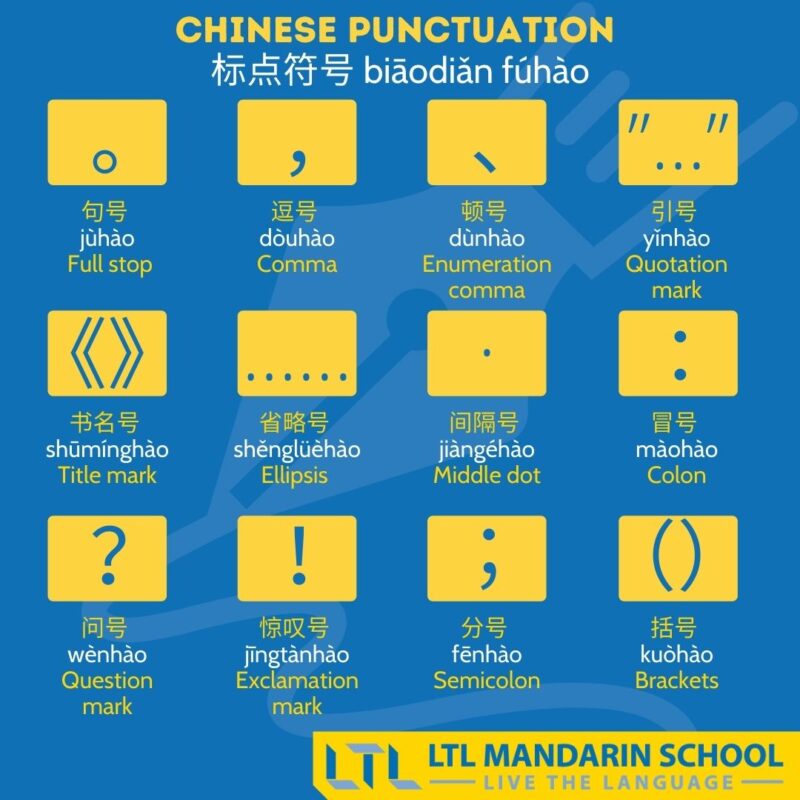}
    \caption{12 Most Common Chinese Punctuation Points}
    \label{fig:punc}
\end{figure}

\begin{figure}[h]
    \centering
    \includegraphics[width=0.4\textwidth]{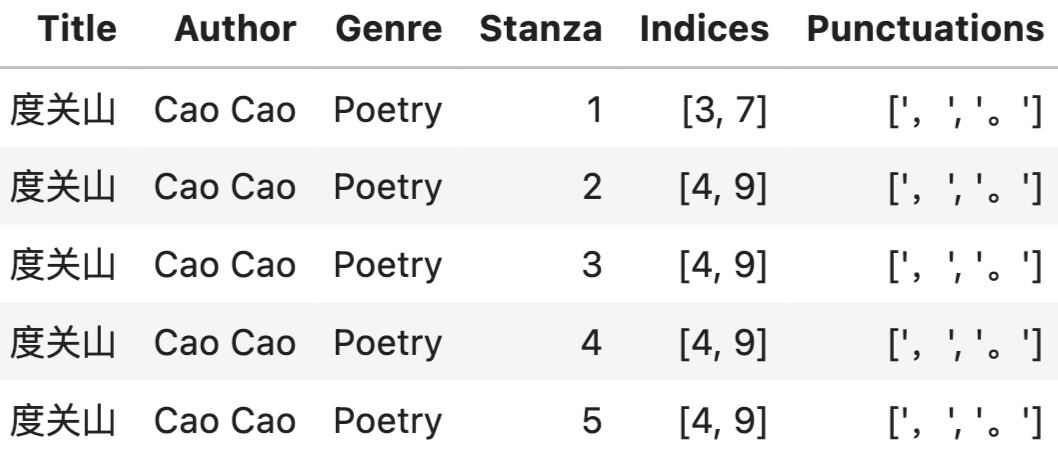}
    \caption{Meta Data}
    \label{fig:metadata}
\end{figure}

\begin{figure}[h]
    \centering
    \includegraphics[width=0.4\textwidth]{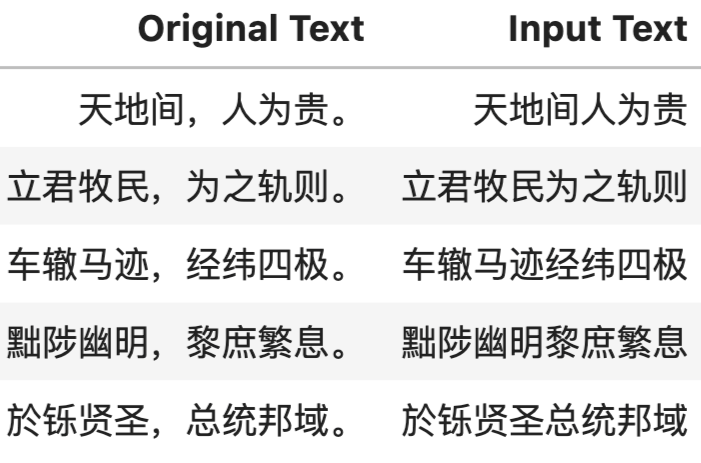}
    \caption{Model Input}
    \label{fig:input}
\end{figure}

\begin{figure}[h]
    \centering
    \includegraphics[width=0.4\textwidth]{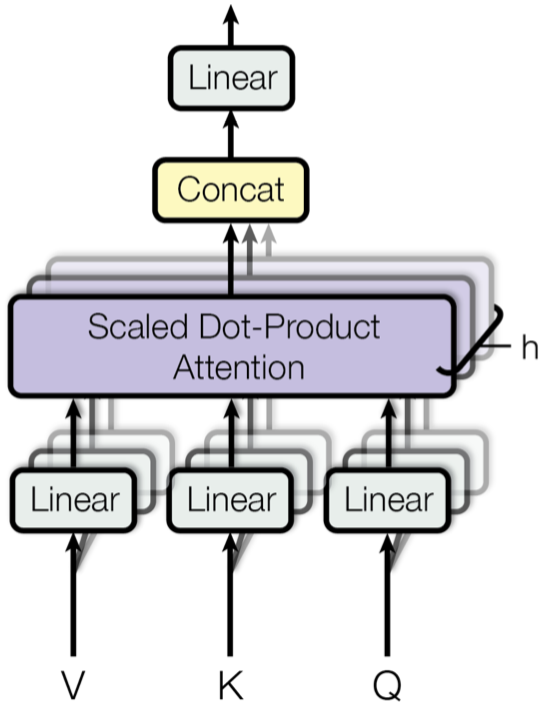}
    \caption{Multi-Head Attention}
    \label{fig:local}
\end{figure}

\begin{figure}[h]
    \centering
    \includegraphics[width=0.4\textwidth]{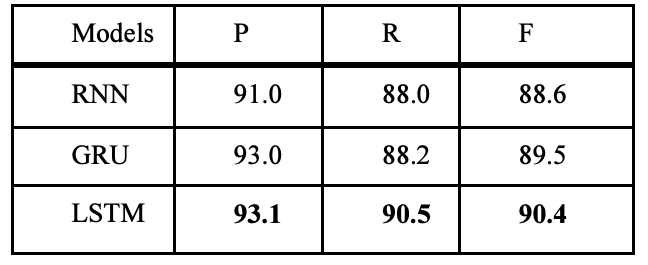}
    \caption{Oh et al's Model Performance with Varying RNN Cells}
    \label{fig:oh}
\end{figure}

\begin{figure}[h]
    \centering
    \includegraphics[width=0.4\textwidth]{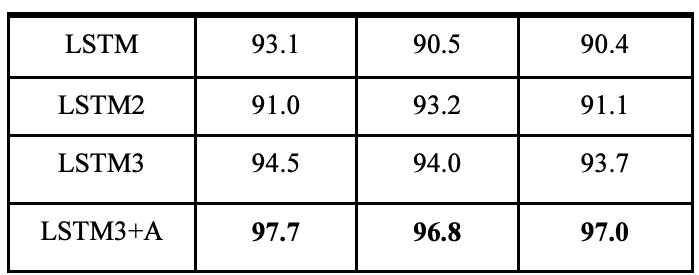}
    \caption{Oh et al's LSTM Performance with Varying Layers and Attention}
    \label{fig:LSTM}
\end{figure}

\begin{figure}[h]
    \centering
    \includegraphics[width=0.4\textwidth]{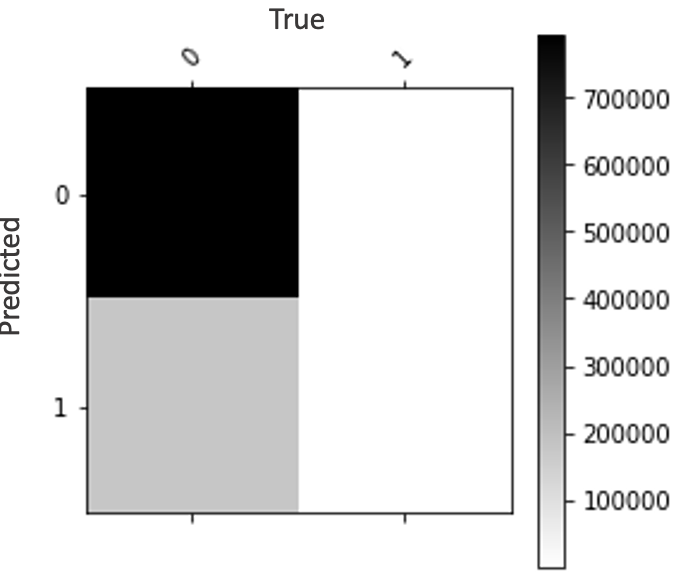}
    \caption{Confusion Matrix of General Punctuation}
    \label{fig:num12}
\end{figure}

\begin{figure}[h]
    \centering
    \includegraphics[width=0.4\textwidth]{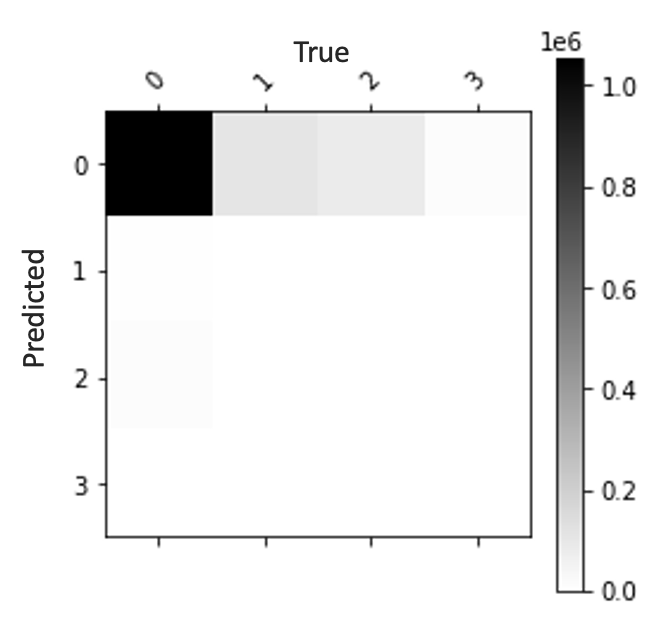}
    \caption{Confusion Matrix of Class-Specific Punctuation}
    \label{fig:num13}
\end{figure}

\end{document}